\documentclass[journal ]{new-aiaa}
\usepackage[utf8]{inputenc}
\usepackage{textcomp}
\usepackage{bm}

\usepackage{graphicx}
\usepackage[version=4]{mhchem}
\usepackage{siunitx}
\usepackage{longtable,tabularx}
\usepackage{multicol}
\usepackage{multirow}
\usepackage{booktabs}

\usepackage{svg}
\usepackage{amsmath}
\usepackage{amsfonts}

\usepackage{mathtools}
\usepackage{mathrsfs}
\usepackage{cleveref}

\usepackage{authblk}

\title{Physics-informed Gaussian Processes for \\ Safe Envelope Expansion}

\author[* {}\textsuperscript{\textdagger}]{D. Isaiah Harp}
\author[* {}\textsuperscript{\ddag}]{Joshua Ott}
\author[{}\textsuperscript{\ddag}]{Dylan M. Asmar}
\author[{}\textsuperscript{\ddag}]{John Alora}
\author[{}\textsuperscript{\ddag}]{Mykel J. Kochenderfer}

\affil[*]{Equal contribution}
\affil[{}\textsuperscript{\textdagger}]{United States Air Force Academy. \texttt{daniel.harp@afacademy.af.edu}}
\affil[{}\textsuperscript{\ddag}]{Stanford University \texttt{\{joshuaott, asmar, jjalora, mykel\}@stanford.edu}}

\begin{document}

\maketitle

\vspace{-5mm}
\begin{abstract}
Flight test analysis often requires predefined test points with arbitrarily tight tolerances, leading to extensive and resource-intensive experimental campaigns. To address this challenge, we propose a novel approach to flight test analysis using Gaussian processes (GPs) with physics-informed mean functions to estimate aerodynamic quantities from arbitrary flight test data, validated using real T-38 aircraft data collected in collaboration with the United States Air Force Test Pilot School. We demonstrate our method by estimating the pitching moment coefficient $ C_m $ without requiring predefined or repeated flight test points, significantly reducing the need for extensive experimental campaigns. Our approach incorporates aerodynamic models as priors within the GP framework, enhancing predictive accuracy across diverse flight conditions and providing robust uncertainty quantification. Key contributions include the integration of physics-based priors in a probabilistic model, which allows for precise computation from arbitrary flight test maneuvers, and the demonstration of our method capturing relevant dynamic characteristics such as short-period mode behavior. The proposed framework offers a scalable and generalizable solution for efficient data-driven flight test analysis and is able to accurately predict the short period frequency and damping for the T-38 across several Mach and dynamic pressure profiles.\footnote[4]{Code and data available at: \url{https://github.com/josh0tt/PIGPSEE}}
\end{abstract}

\section*{Nomenclature}
{\renewcommand\arraystretch{1.0}
\noindent\begin{longtable*}{@{}l @{\quad=\quad} l@{}}
$a$ & regression coefficient \\
$\alpha$ & angle of attack (radians) \\
$\alpha_{\text{trim}}$ & trim angle of attack (radians) \\
$b$ & regression coefficient \\
$\bar{q}$ & dynamic pressure (lb/ft$^2$) \\
$c$ & regression coefficient \\
$C_m$ & pitching moment coefficient \\
$C_{m_\alpha}$ & derivative of $C_m$ with respect to $\alpha$ \\
$C_{m_Q}$ & derivative of $C_m$ with respect to $Q$ \\
$C_{m_{\delta_e}}$ & derivative of $C_m$ with respect to $\delta_e$ \\
$d$ & regression coefficient \\
$\delta_e$ & stabilator deflection (radians) \\
$\delta_{e_{\text{trim}}}$ & trim stabilator deflection (radians) \\
$f$ & true function \\
$k(\mathbf{x}, \mathbf{x}')$ & covariance function (kernel) \\
$M$ & Mach number \\
$M_{\alpha}$ & derivative of pitching moment with respect to $\alpha$ \\
$M_Q$ & derivative of pitching moment with respect to $Q$  \\
$P$ & roll rate (rad/s) \\
$Q$ & pitch rate (rad/s) \\
$R$ & yaw rate (rad/s) \\
$\rho$ & air density (slugs/ft$^3$) \\
$U_1$ & aircraft trim speed along the \textit{x}-axis (ft/s) \\
$X$ & observed inputs \\
$X^*$ & test inputs \\
$y$ & noisy observations \\
$\hat{y}$ & predicted outputs \\
$Z_{\alpha}$ & derivative of \textit{Z}-force with respect to $\alpha$ \\
$z$ & zero-mean Gaussian noise \\
$\zeta_{SP}$ & short period damping ratio \\
$\omega_{SP}$ & short period frequency (Hz) \\
$\nu$ & variance of noise \\
$\mathbf{x}^{(i)}$ & $i$th observed state vector \\
$\mu^*$ & predictive mean \\
$\Sigma^*$ & predictive covariance \\
$m(\mathbf{x})$ & mean function \\
$\mathbf{m}(X)$ & vector of mean functions evaluated at $X$ \\
$\mathbf{K}(X, X')$ & covariance matrix \\
$\mathbf{k}(\mathbf{x}, \mathbf{X})$ & covariance vector \\
\end{longtable*}}

\section{Introduction}\label{sec:intro}

A central challenge in flight testing is accurately determining key aerodynamic quantities without exhaustive, repetitive data collection \cite{morelli2015global, morelli2023sysid, yechout2003introduction}. Traditional approaches often rely on extensive flight test campaigns, in which a large number of test points with narrow tolerances are flown across an aircraft envelope \cite{yechout2003introduction, dtic1961t38}. In one major Air Force envelope expansion program with 2,199 test points, pilots flew 13,500 maneuvers. On average, each maneuver had to be repeated more than six times to achieve data tolerances and quality \cite{wendy2023rpg}. Current flight test processes are time-consuming and labor-intensive, leading to thousands of hours of repeated testing. In this work, we introduce an approach that uses Gaussian processes (GPs) with a physics-informed mean function to calculate aerodynamic quantities, specifically the pitching moment coefficient $ C_m $, from arbitrary flight test data. While $ C_m $ serves here as a demonstrative case, the broader contribution lies in our approach, which offers a generalizable method to eliminate redundant test points and streamline the flight test process.

The implications of this approach extend beyond modeling $ C_m $ under varying conditions; they suggest a transformative way to approach flight testing as a whole. By anchoring a GP in aerodynamic principles through its mean function, our method enables precise estimation of $ C_m $ across diverse aircraft states without the need for meticulously predefined data points, while simultaneously relieving the test pilot of the burden to achieve the parameters of those points. This capability not only reduces the experimental burden but also allows test pilots and engineers to operate more flexibly, gathering data without adhering to rigid tolerances \cite{havekirby, havepalpatine}. Consequently, this framework has the potential to substantially reduce the time required to conduct a flight test campaign by at least a factor of six \cite{wendy2023rpg}.

The challenge of creating a generalizable framework for flight test data analysis arises from the inherent variability and non-linearity in flight dynamics \cite{yechout2003introduction}. Standard data-driven models, though useful, require repeated measurements across a spectrum of flight conditions and are constrained by the structure of available data \cite{morelli1995mof, morelli2015global}. Furthermore, traditional methods struggle with extrapolation and often fail to incorporate the physical constraints inherent to the fundamentals of known aerodynamic relationships \cite{ljung1998system}. A purely data-driven approach to estimating $ C_m $ or similar quantities could quickly become unreliable without embedding prior aerodynamic knowledge, especially under complex flight conditions \cite{brunton2022data}.

Previous methods for calculating $ C_m $ have often focused on fixed-condition models or demanded intensive data collection strategies, limiting their flexibility in real-world applications \cite{havekirby, havepalpatine}. Our approach departs from these by integrating physics-based aerodynamic models directly into the GP framework as priors. This integration allows the model to leverage the known aerodynamic relationships while refining its predictions with new data. Unlike traditional methods that require narrow tolerances on test conditions, our approach enables the calculation of $ C_m $ from diverse and arbitrary flight maneuvers, fundamentally transforming the way flight test data can be gathered and analyzed.

This paper presents a Gaussian process model, informed by aerodynamic principles, to demonstrate the feasibility of estimating $ C_m $ from arbitrary flight test data without rigid data requirements. We apply our method to historical flight test data, illustrating its effectiveness in capturing short-period dynamics relevant to $ C_m $. We also discuss the key limitations of our approach, including the reliance on training data to refine predictions and the assumption of steady-state conditions for stability analysis. We demonstrate that our method represents a powerful, flexible framework that can generalize to other aerodynamic quantities. This physics-informed GP approach has the potential to streamline the flight test process, reduce costs, and expand the applicability of physics-informed data-driven models in aerospace engineering and especially flight test engineering.

\section{Related Work}\label{sec:related_work}
Traditional flight test methods rely heavily on structured test plans that require predefined maneuvers with narrow tolerances and extensive data collection to capture the dynamics of the aircraft under various condition \cite{dtic1961t38}. Classic system identification approaches have been instrumental in developing transfer functions and stability derivatives through carefully controlled maneuvers \cite{shepherd2010limited, welborn2010tfs, ljung1998system, tangirala2018principles, keesman2011system, kostelich1992problems, vandenberghe2012convex}. These techniques, however, depend on a dense sampling of flight conditions, with fixed-point data collection that often incurs significant operational costs due to repeated testing \cite{yechout2003introduction}. Moreover, they struggle to disambiguate open-loop from closed-loop response \cite{knapp2018vista}.

Recent advances in data-driven methods, particularly machine learning models, have enabled researchers to approach flight test analysis with greater flexibility \cite{brunton2016discovering, kaiser2018sparse, vincent2010input, noel2017nonlinear, tsiamis2022learning, lee2024active, brunton2022data}. Gaussian processes have emerged as a powerful tool for capturing non-linear relationships with built-in uncertainty quantification \cite{OTT2024104814, kochenderfer2019algorithms, williams2006gaussian, ott2024input}. Gaussian processes have been applied in aerospace to estimate aerodynamic forces and moments, benefiting from the probabilistic structure that allows for robust prediction in uncertain environments \cite{kumar2019gpr}. Yet, despite their advantages, these models often lack the capacity to extrapolate meaningfully without substantial data, which can lead to unreliable predictions in sparse or unexplored regions of the flight envelope \cite{kochenderfer2019algorithms}.

A significant body of work has focused on enhancing Gaussian processes through the integration of physical insights, often referred to as physics-informed Gaussian processes \cite{pang2020physics, yang2019physics, tartakovsky2023physics}. More generally, physics-informed priors have been used to embed known behaviors into the model structure, improving generalization to unseen conditions while reducing data requirements \cite{raissi2019physics}. In the context of aerospace applications, incorporating aerodynamic relationships as priors can be particularly beneficial, providing a mechanism to guide the model's behavior based on foundational aerodynamic principles \cite{petersen2022wind, cross2024spectrum}. While such physics-informed models have shown promise in domains like fluid flow and structural dynamics, their application to flight testing, particularly in estimating stability derivatives directly from non-standardized flight data, remains relatively unexplored.

Our approach builds on these advancements by combining Gaussian processes with aerodynamic priors derived from wind tunnel or existing flight test data. Unlike traditional system identification methods or standard Gaussian process models, our framework uses a physics-informed mean function constructed from known aerodynamic relationships to calculate the pitching moment coefficient $C_m$ without requiring predefined test maneuvers. This approach aligns with recent trends toward hybrid modeling, where data-driven models are enriched with physical knowledge to enhance predictive performance and reduce the experimental footprint \cite{brunton2016discovering, brunton2022data, kaiser2018sparse}. Through this integration, we address the limitations of purely data-driven or traditional system identification methods, achieving robust, scalable predictions from arbitrary test data \cite{karniadakis2021physics}. This contribution not only enables more flexible flight test strategies but also represents a shift toward more efficient, cost-effective approaches in aircraft envelope expansion.

\section{Background}\label{sec:background}
This article uses Gaussian process notation consistent with that of \citeauthor{kochenderfer2019algorithms} \cite{kochenderfer2019algorithms}. A Gaussian process defines a distribution over functions. A Gaussian process is parameterized by a mean function $m$ and a kernel function $k$. The mean function encodes prior knowledge about the function’s behavior, and the kernel dictates the smoothness and other properties of the functions. Gaussian processes can model noisy observations of the true function $f$ that we are interested in predicting. These noisy observations are given by $y = f(\mathbf{x}) + z$ where $f$ is deterministic but $z$ is zero-mean Gaussian noise $z \sim \mathcal{N}(0, \nu)$. If we already have a set of observed points $X$ and the corresponding $\mathbf{y}$, we can predict the values $\mathbf{\hat{y}}$ at points $X^*$. The joint distribution is given by: \begin{equation}
\begin{bmatrix}
    \mathbf{\hat{y}} \\
    \mathbf{y}
\end{bmatrix}
\sim \mathcal{N} \left( 
    \begin{bmatrix}
    \mathbf{m}(X^*) \\
    \mathbf{m}(X)
    \end{bmatrix}, 
    \begin{bmatrix}
    \mathbf{K}(X^*, X^*) & \mathbf{K}(X^*, X) \\
    \mathbf{K}(X, X^*) & \mathbf{K}(X, X) + \nu \mathbf{I}
    \end{bmatrix} 
\right)
\end{equation} with conditional distribution $\mathbf{\hat{y}} \mid \mathbf{y}, \bm{\nu} \sim \mathcal{N}(\bm{\mu^*}, \bm{\Sigma^*})$ where: \begin{equation}
\begin{aligned}
    \bm{\mu^*} &= \mathbf{m}(X^*) + \mathbf{K}(X^*, X)(\mathbf{K}(X, X) + \nu \mathbf{I})^{-1}(\mathbf{y} - \mathbf{m}(X)) \\
    \bm{\Sigma^*} &= \mathbf{K}(X^*, X^*) - \mathbf{K}(X^*, X)(\mathbf{K}(X, X) + \nu \mathbf{I})^{-1} \mathbf{K}(X, X^*). \label{eq:gp_mean_and_covar}
\end{aligned}
\end{equation} In the equations above, we use the functions $\mathbf{m}$ and $\mathbf{K}$ which for any finite set of points $X = \{x^{(1)}, \dots, x^{(n)}  \}$ and $X' = \{x'^{(1)}, \dots, x'^{(m)}  \}$ are given by: \begin{equation}
    \begin{aligned}
        \mathbf{m}(X) &= \begin{bmatrix} m(x^{(1)}), \dots, m(x^{(n)}) \end{bmatrix} \\
        \mathbf{K}(X, X') &= 
        \begin{bmatrix}
        k(x^{(1)}, x'^{(1)}) & \dots & k(x^{(1)}, x'^{(m)}) \\
        \vdots & \ddots & \vdots \\
        k(x^{(n)}, x'^{(1)}) & \dots & k(x^{(n)}, x'^{(m)})
        \end{bmatrix}.
    \end{aligned}
\end{equation} %

\section{Methods}\label{sec:methods}
We use a Gaussian processes to model the mapping from aircraft state variables to aerodynamic quantities of interest, specifically the pitching moment coefficient \( C_m \). The state vector \( \mathbf{x} \) is given by: 
\begin{equation}
    \mathbf{x}^{(i)} = \left[M, \rho, \bar{q}, P, Q, R, \alpha, \delta_e\right]^{\top} \label{eq:state_vector},
\end{equation} where \( M \) is the Mach number, \( \rho \) is the air density (computed from the ideal gas law using pressure altitude and outside air temperature), \( \bar{q} \) is the dynamic pressure, \( P, Q, R \) are the roll, pitch, and yaw rates, \( \alpha \) is the angle of attack, and \( \delta_e \) is the stabilator deflection. It is important to note that orientation variables, such as pitch angle or roll angle, are not part of the state vector because $C_m$ depends primarily on rates and other flight parameters, rather than orientation itself. However, orientation state variables (and indeed, any variable of interest to the experimenter) can be added to the state vector in general. %

Our objective is to learn the underlying function $C_m = f(\mathbf{x})$ that maps the state variables to the pitching moment coefficient. To achieve this, we construct a Gaussian process over \( f \) and update it using observed data \( \{ (\mathbf{x}^{(i)}, y^{(i)}) \}_{i=1}^N \), where \( y^{(i)} \) is a noisy observation of \( C_m \) at state $\mathbf{x}^{(i)}$.

We incorporate physics-based priors into the mean function \( m(\mathbf{x}) \) of the Gaussian process, reflecting established aerodynamic relationships. Specifically, we use the mean function derived from the Morelli aerodynamic model \cite{morelli2015global}: \begin{equation} 
    C_{m} = 
    \theta_{29} + \theta_{30} \alpha + \theta_{31} Q + \theta_{32} \delta_e + \theta_{33} \alpha Q + \theta_{34} \alpha^2 Q + \theta_{35} \alpha^2 \delta_e + \theta_{36} \alpha^3 Q + \theta_{37} \alpha^3 \delta_e + \theta_{38} \alpha^4 %
    \label{eq:mean-function}.
\end{equation} The Morelli model was empirically determined from wind tunnel experiments to extract the dominant regressor coefficients upon which $C_m$ is known to depend \cite{morelli2015global}. We use the coefficients determined by Morelli for the A-7E Corsair II. However, all flight test data collection and experiments discussed in \cref{sec:results} were conducted using the T-38C. We intentionally use a prior aerodynamic model that was known to be incorrect. While of a similar generation of aircraft, the A-7 is substantially larger than the T-38C and differs in stability and control characteristics \cite{morelli2015global, shepherd2010limited}. For our analysis, we aim to start with a similar, yet fundamentally flawed prior, allowing the physics-informed, data-driven method to resolve the discrepancies. The A-7 mean function encodes our prior knowledge about how \( C_m \) depends on \( \alpha \), \( Q \), and \( \delta_e \), allowing the Gaussian process to focus on modeling deviations from this baseline resulting from the particular shape and size of the T-38C and its aerodynamic surfaces.

The covariance function $k(\mathbf{x}, \mathbf{x}')$ determines the smoothness and generalization properties of the Gaussian process. We used the neural network kernel: 

\begin{equation}
k(\mathbf{x}, \mathbf{x}') = \sin^{-1} \left( \frac{\mathbf{x}^\top \mathbf{x}'}{\sqrt{(1 + \|\mathbf{x}\|_2^2 / 2)(1 + \|\mathbf{x}'\|_2^2 / 2)}} \right).
\end{equation} This kernel captures complex, non-linear relationships between inputs and outputs \cite{williams2006gaussian, hornik1993some, kochenderfer2019algorithms}.

After training, the Gaussian process provides a posterior distribution over functions, enabling us to make predictions about \( C_m \) at new state vectors \( \mathbf{x}^* \). The predictive mean and covariance are given by \cref{eq:gp_mean_and_covar}.

\begin{figure}
    \centering
    \includegraphics[width = 0.85\linewidth]{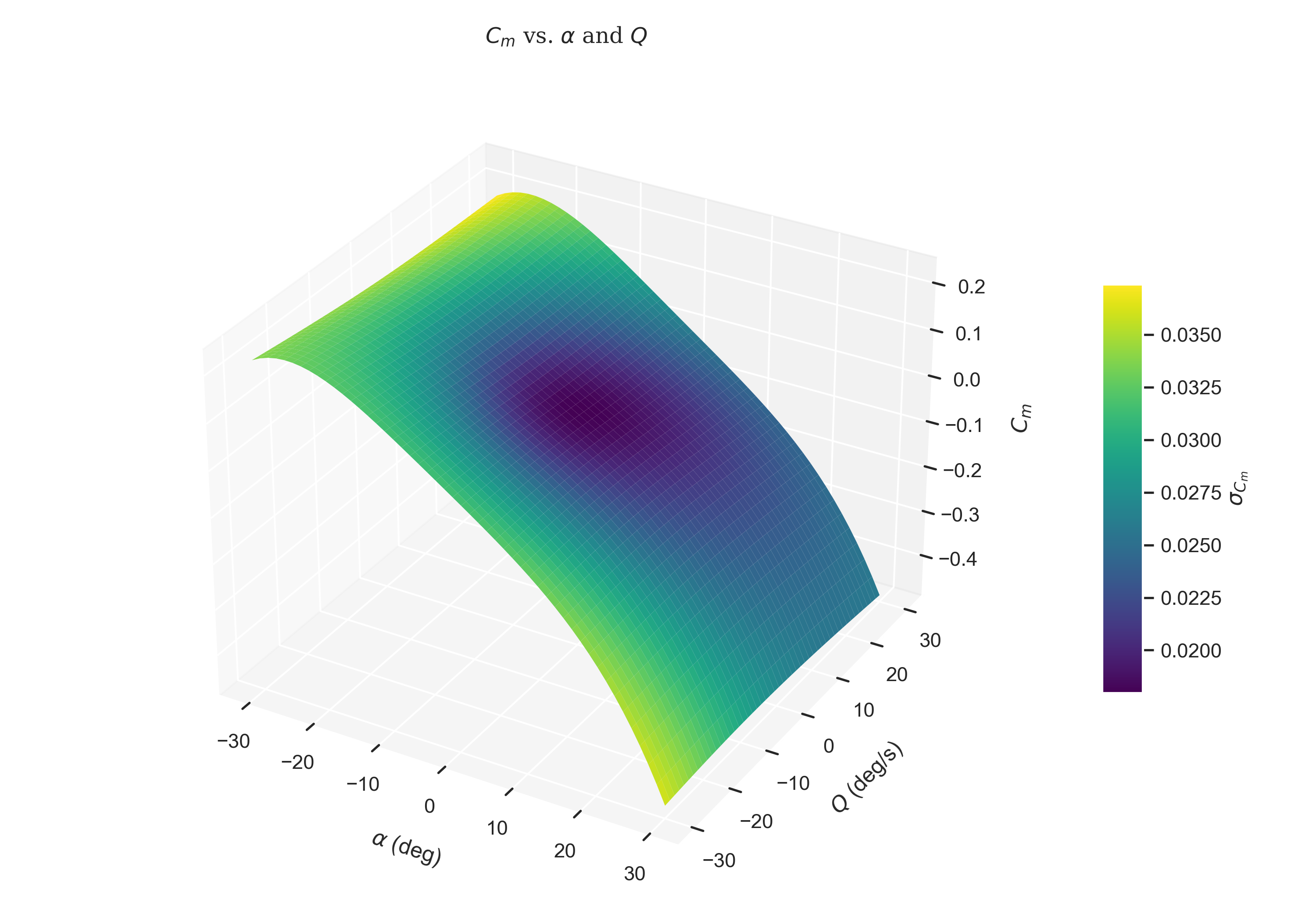}
    \caption{Surface plot showing the relationship between \( \alpha \), \( Q \), and \( C_m \) learned by the Gaussian process. The surface is colored by the model's uncertainty values.}
    \label{fig:surface_plot}
\end{figure}

An example of a surface extracted from the Gaussian process is shown in \cref{fig:surface_plot}. This 3D surface illustrates how variations in $\alpha$ and $Q$ impact the pitching moment coefficient, with all other state variables (the remaining dimensions of the hypersurface) fixed at what we define as trim values. The process of determining these trim values will be described in the following section. Observe that the Gaussian process also provides uncertainty estimates. The model is most certain at moderate values of $\alpha$ and $Q$, as we would expect.

\subsection{Prediction of the Short Period} \label{gp-sp}
One of the key advantages of Gaussian processes is that they provide a smooth, differentiable surface from the posterior estimate. This allows us to compute classical stability derivatives by differentiating the Gaussian process at specific combinations of the state variables representing specific flight conditions.

\subsubsection{Trim Conditions and Stability Derivatives} \label{trim}

Since the domain of our surface is unbounded, we can compute derivatives at any point, including those corresponding to non-physical combinations of state variables. %
To reproduce the aircraft's true dynamical parameters and compare them to historical databases, we collapsed the surface to specific, realistic combinations of state variables that we called \emph{trim conditions}. In our usage, a trim condition is a steady-state flight condition at some particular altitude and airspeed where the aircraft is in equilibrium and the rate components of the state vector are zero. In order to ensure physical realism at the trim condition we use to differentiate the Gaussian process, we had to experimentally determine the other components of the state vector (\textit{i.e.}, $\alpha$ and $\delta_{e}$) that actually produce such a condition in free flight. 

We experimentally determine trim functions by performing \emph{trim shots} at different altitudes and airspeeds. During a trim shot, the aircraft is flown in steady, unaccelerated flight and the corresponding state variables are recorded. We regress the data to obtain empirical relationships for the trim angle of attack \( \alpha_{\text{trim}} \) and the trim stabilator deflection \( \delta_{e_{\text{trim}}} \) as functions of dynamic pressure \( \bar{q} \): \begin{align}
\alpha_{\text{trim}}(\bar{q}) &= a \cdot \exp(-b \bar{q}), \label{trim-aoa} \\
\delta_{e_{\text{trim}}}(\bar{q}) &= c + d \cdot \ln(\bar{q}), \label{trim-stab}
\end{align} where \( a, b, c, \) and \( d \) are coefficients obtained from the regression of experimental data. To account for Mach effects, we calculate multiple trim functions for multiple Mach ranges.

These relationships allow us to define the trim state \( \mathbf{x}_{\text{trim}} \) at any given dynamic pressure. %
We compute the stability derivatives at these trim states to ensure our results align with physical realism and to facilitate meaningful comparisons with historical data. However, it is important to note that we specify zero rates and particular altitude and airspeed combinations only to compare with historical flight test data collected from straight-and-level, trimmed flight. One key advantage of our approach is that we can query the Gaussian process at any flight condition. Provided the state variables we supply are physically achievable, the surface will yield a meaningful prediction for that flight condition.

\subsubsection{Differentiation of the Gaussian Process} \label{ss:differentiation}

The stability derivatives of $C_{m}$ with respect to $\alpha$ and $Q$ are $C_{m_\alpha}$ and $C_{m_Q}$ respectively. The stability derivatives are obtained by differentiating the Gaussian process posterior mean function \( \mu(\mathbf{x}) \) with respect to the relevant state variables at the trim condition \( \mathbf{x}_{\text{trim}} \): \begin{align}
C_{m_\alpha} &\equiv \left. \frac{\partial \mu(\mathbf{x})}{\partial \alpha} \right|_{\mathbf{x} = \mathbf{x}_{\text{trim}}}, \label{Cm_alpha} \\
C_{m_Q} &\equiv \left. \frac{\partial \mu(\mathbf{x})}{\partial Q} \right|_{\mathbf{x} = \mathbf{x}_{\text{trim}}}.  \label{Cm_Q}
\end{align} The posterior mean function \( \mu(\mathbf{x}) \) is: \begin{equation}
\mu(\mathbf{x}) = m(\mathbf{x}) + \mathbf{K}(\mathbf{x}, X)^\top \mathbf{A},
\end{equation} with \begin{equation}
\mathbf{A} = (\mathbf{K}(X,X) + \nu \mathbf{I})^{-1} (\mathbf{y} - \mathbf{m}(X)).
\end{equation} The gradient of \( \mu(\mathbf{x}) \) with respect to \( \mathbf{x} \) is: \begin{equation}
\nabla_{\mathbf{x}} \mu(\mathbf{x}) = \nabla_{\mathbf{x}} m(\mathbf{x}) + \left( \nabla_{\mathbf{x}} \mathbf{K}(\mathbf{x}, X) \right)^\top \mathbf{A}. \label{eq:grad_gp_mean}
\end{equation} We used automatic differentiation tools to compute the gradients of \cref{eq:grad_gp_mean} efficiently.

\subsubsection{Calculation of Short Period Dynamics}

With the stability derivatives \( C_{m_\alpha} \) and \( C_{m_Q} \) obtained from the Gaussian process, we calculate the short period natural frequency \( \omega_{n_{sp}} \) and damping ratio \( \zeta_{sp} \), which are useful parameters for understanding the longitudinal dynamic response.

The short period dynamics are derived from the linearized longitudinal equations of motion, following a two degree-of-freedom approximation \cite{yechout2003introduction}. Under these approximations, the short period frequency and damping relate to the dimensional stability derivatives $M_Q$, $M_{\alpha}$, $M_{\dot{\alpha}}$, and $Z_{\alpha}$. The dimensional derivatives of the moment $M$ relate to the non-dimensional derivatives of $C_{M}$ calculated from the Gaussian process, and are provided by \citeauthor{nelson1998stability} \cite{nelson1998stability}. Similarly, the dimensional derivative \(Z_{\alpha}\) is determined by fitting an independent Gaussian process model to map the same state variables to the aerodynamic force coefficient \(C_Z\). The derivative \(Z_{\alpha}\) is then computed by differentiating this Gaussian process model following the same procedure used for the moment derivatives.  %

The expressions for \( \omega_{n_{sp}} \) and \( \zeta_{sp} \) are then: \begin{align}
\omega_{SP} &= \sqrt{ \frac{ -Z_{\alpha} M_Q }{ U_1 } - M_{\alpha} }, \label{eq:omega-gp} \\
\zeta_{SP} &= - \frac{ M_Q + \left( M_Q / 3 \right) + (Z_{\alpha} / U_1) }{ 2 \omega_{SP} }, \label{eq:zeta-gp}
\end{align} except in \cref{eq:zeta-gp} we replace $M_{\dot{\alpha}}$ with $\frac{1}{3}M_Q$ as proposed by \citeauthor{yechout2003introduction} \cite{yechout2003introduction}.

\begin{table}[t]
\centering
\begin{tabular}{
    ll
    S[table-format=5,group-separator={}]
    S[table-format=1.2]
    S[table-format=4,group-separator={}]
    S[table-format=1.2]
    S[table-format=1.2]
    c
}
\toprule
\multirow{2}{*}{Reference} & \multirow{2}{*}{Data Source} & \multicolumn{1}{c}{Altitude} & \multicolumn{1}{c}{Mach} & \multicolumn{1}{c}{q} & \multicolumn{1}{c}{$\omega_{sp}$} & \multicolumn{1}{c}{$\zeta_{sp}$} & \multirow{2}{*}{Symbol} \\
& & \multicolumn{1}{c}{[ft]} & \multicolumn{1}{c}{\ } & \multicolumn{1}{c}{[lb/ft\textsuperscript{2}]} & \multicolumn{1}{c}{[Hz]} & \multicolumn{1}{c}{\ } & \\
\midrule
\multirow{4}{*}{\citeauthor{shepherd2010limited} \cite{shepherd2010limited}} & \multirow{4}{*}{SIDPAC} & \num{10356} & \num{0.71} & \num{513} & \num{0.72} & \num{0.32} & \multirow{4}{*}{$\blacktriangle$} \\
& & \num{31753} & \num{0.71} & \num{203} & \num{0.38} & \num{29} & \\
& & \num{32460} & \num{1.08} & \num{458} & \num{0.90} & \num{0.22} & \\
& & \num{22751} & \num{0.91} & \num{498} & \num{0.86} & \num{0.32} & \\
\midrule
\multirow{4}{*}{\citeauthor{welborn2010tfs} \cite{welborn2010tfs}} & \multirow{4}{*}{CIFER} & \num{20000} & \num{0.90} & \num{551} & \num{0.64} & \num{0.36} & \multirow{4}{*}{$\blacksquare$} \\
& & \num{20000} & \num{0.70} & \num{334} & \num{0.46} & \num{0.31} & \\
& & \num{20000} & \num{0.50} & \num{170} & \num{0.34} & \num{0.31} & \\
& & \num{20000} & \num{0.90} & \num{551} & \num{0.66} & \num{0.38} & \\
& & \num{20000} & \num{0.70} & \num{334} & \num{0.47} & \num{0.34} & \\
& & \num{20000} & \num{0.50} & \num{170} & \num{0.34} & \num{0.35} & \\

\midrule
\multirow{2}{*}{TPS Student} & \multirow{2}{*}{Experimental-Doublet} & \num{20000} & \num{0.90} & \num{551} & \num{0.65} & \num{0.33} & \multirow{2}{*}{$\bullet$} \\
& & \num{20000} & \num{0.70} & \num{334} & \num{0.49} & \num{0.31} & \\
\midrule
\multirow{7}{*}{AFFTC} & \multirow{7}{*}{Experimental-Various} & \num{10000} & \num{0.35} & \num{125} & \num{0.37} & \num{0.30} & \multirow{7}{*}{$\blacklozenge$} \\
& & \num{10000} & \num{0.35} & \num{125} & \num{0.37} & \num{0.23} & \\
& & \num{25000} & \num{0.60} & \num{198} & \num{0.36} & \num{0.27} & \\
& & \num{45000} & \num{0.81} & \num{141} & \num{0.38} & \num{0.26} & \\
& & \num{10000} & \num{0.89} & \num{807} & \num{0.83} & \num{0.35} & \\
& & \num{45000} & \num{0.92} & \num{182} & \num{0.42} & \num{0.18} & \\
& & \num{25000} & \num{0.90} & \num{445} & \num{0.67} & \num{0.30} & \\
& & \num{45000} & \num{1.06} & \num{242} & \num{0.56} & \num{0.13} & \\
& & \num{25000} & \num{1.06} & \num{653} & \num{0.83} & \num{0.17} & \\
\bottomrule
\end{tabular}
\caption{The consolidated short period data collected for comparison. The symbols cross-reference with \cref{fig:comparison-points}.} \label{table:SP-table}
\end{table}

\section{Results}\label{sec:results}
To assess the accuracy of our methods, we gathered historical comparison data on the T-38 short period mode from four sources, encompassing a range of methods.

\begin{figure}[t]
    \centering
    \includegraphics[width = \linewidth]{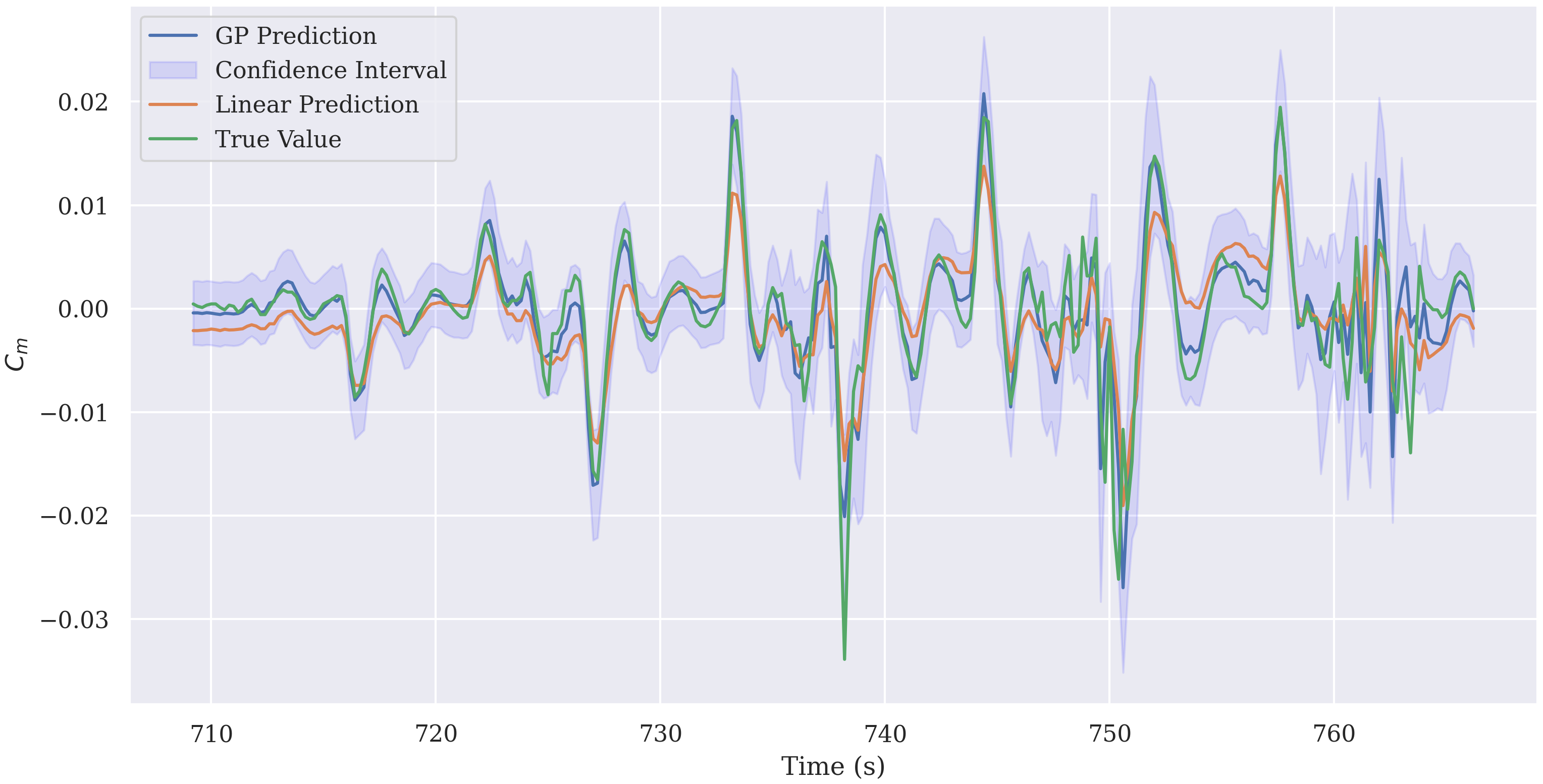}
    \caption{$C_m$ prediction using our Gaussian process based approach. The linear model is constructed using linear regression between the state vector given in  \cref{eq:state_vector} and $C_m$.}
    \label{fig:cm_pred}
\end{figure}

\begin{enumerate}
    \item A system identification investigation flown by the United States Air Force Test Pilot School (TPS) in the T-38A\footnote{We do not expect significant differences in the short period between the T-38A and the T-38C.}. This investigation used frequency sweeps to excite the short period mode, and the System Identification Programs for AirCraft (SIDPAC) toolbox for MATLAB to identify stability derivatives using a linear model structure \cite{shepherd2010limited}.
    
    \item A system identification approach using the Comprehensive Identification from Frequency Responses (CIFER) software package to identify transfer functions directly. The results included $\frac{\alpha}{\delta_e}$ transfer functions and $\frac{\theta}{\delta_e}$ transfer functions, both of which were used for comparison \cite{welborn2010tfs}.
    
    \item Two T-38 flying qualities reports completed by the Air Force Test Pilot School (TPS). These investigations both used doublets to excite the short period mode, and the amplitudes and phases of the oscillation were calculated directly from the data acquisition system \cite{havekirby, havepalpatine}.
    
    \item The Air Force Flight Test Center (AFFTC) final stability and control report on the T-38A. Various methods were used, as explained by \citeauthor{dtic1961t38} \cite{dtic1961t38}.\footnote{Pitch damper OFF and pitch damper ON stability and control results were presented in the test report. We used the pitch damper OFF values, since the pitch damper was removed for the T-38C.} %
\end{enumerate}

\Cref{table:SP-table} tabulates the collected data. There are some limitations with these comparison points. Neither the Welborn nor the TPS student data report dynamic pressure (nor actual air temperature), and so a standard atmosphere is assumed. Also, the data is not standardized to a common center of gravity (CG) location; it is known that CG location alters the short period frequency. %

\Cref{fig:cm_pred} illustrates the predictive performance of our Gaussian process model for the pitching moment coefficient $ C_m $ during a rollercoaster maneuver conducted at $0.7$M in the T-38C aircraft. The true values of $ C_m $ are calculated by non-dimensionalizing the summation of moments equation \begin{equation}
    I_{y}\dot{Q}+(I_{x}-I_{z})PR+I_{xz}(P^{2}-R^{2}) =\sum M_{y} \equiv \bar{M}. \label{eq:Mbar}
\end{equation} These values are indicated by the dashed line. The solid line denotes the GP prediction and a shaded region captures the 95\% confidence interval.

The GP model successfully captures the general trend of $ C_m $, providing reasonably accurate estimates while also quantifying the inherent uncertainty in dynamic maneuvers. As shown in the figure, the GP prediction closely follows the true moment values throughout the maneuver, with the uncertainty bands encompassing the true values in most regions. \Cref{tab:cm_comparison} also compares the estimates of the dynamic parameters $C_{m_{\alpha}}$, $C_{m_{\delta_{e}}}$, $C_{m_{Q}}$, $\omega_{SP}$, and $\zeta_{SP}$ between our proposed method, the results from \citeauthor{shepherd2010limited} \cite{shepherd2010limited}, and a linear model constructed using linear regression between the state vector given in \cref{eq:state_vector} and $C_m$.

\begin{figure}
    \centering
    \includegraphics[width = 0.85\linewidth]{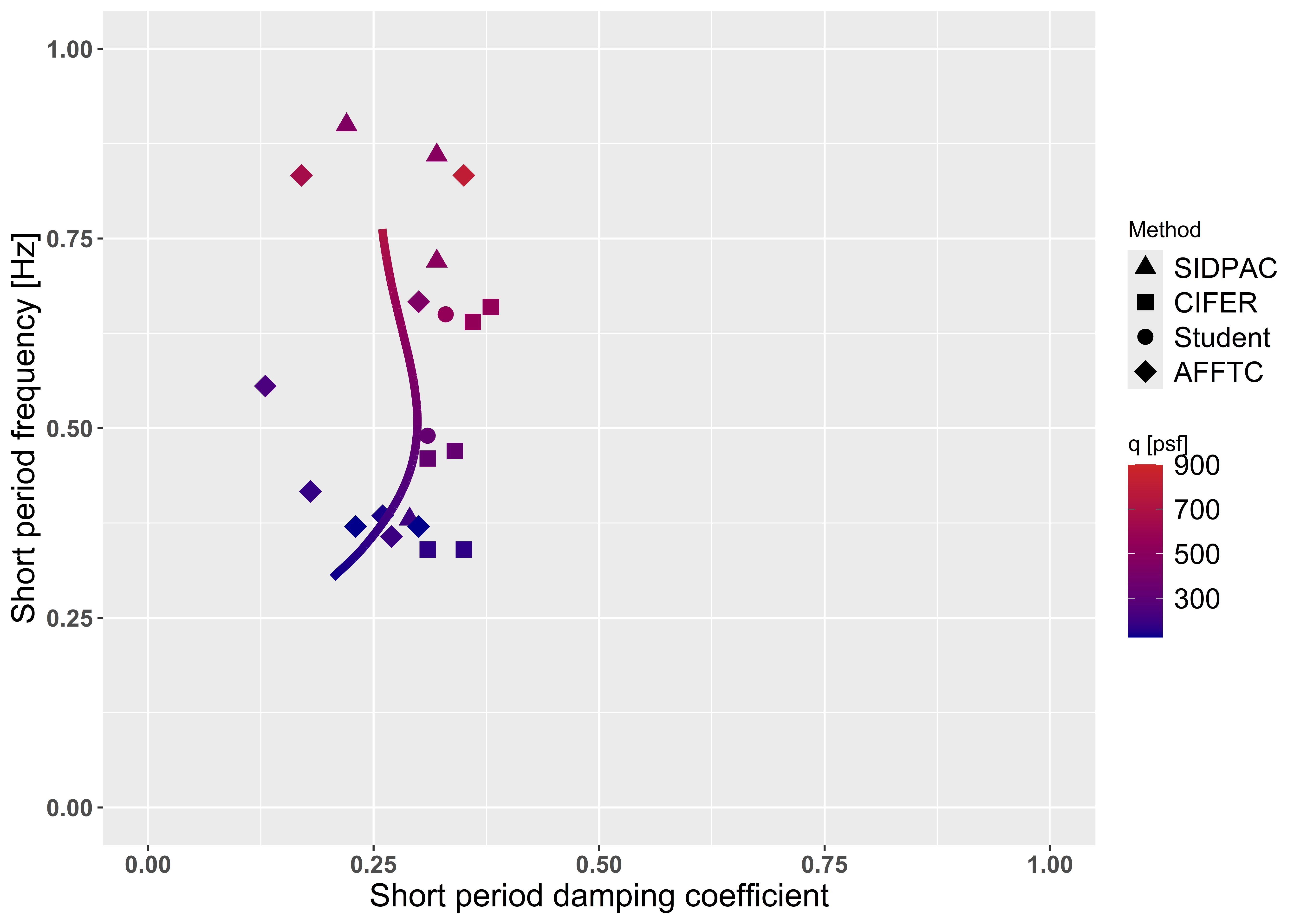}
    \caption{The consolidated short period data collected for comparison, along with continuous predictions produced by our method with varying dynamic pressure (denoted as q in this figure).}
    \label{fig:comparison-points}
\end{figure}

\Cref{fig:comparison-points} shows the complete set of short period frequency and damping coefficients from the historical data provided in \cref{table:SP-table}. As we have emphasized, the short period frequency is a strong implicit function of dynamic pressure. Since our physics-informed data-based architecture outputs a full distribution, we are able to calculate the short period frequency and damping at any dynamic pressure. \Cref{fig:comparison-points} shows a continuous prediction %
using our method for the full range of dynamic pressures appearing in the comparison data. 

The correspondence with the cluster of historical data demonstrates relatively strong agreement. This agreement occurs despite the fact that the roller coaster maneuver used to calculate these short period parameters lacked the high frequency content required by SIDPAC, CIFER, and other system identification methods \cite{shepherd2010limited, welborn2010tfs}. In those methods, quantification of a dynamical response requires excitation of the aircraft at or near the frequencies to be identified. However, in the roller coaster data the pilot input AOA oscillations at a rate of around 1.3 deg/sec, or a mere 0.0036 Hz. This frequency input into the system was just 1\% of the average short period frequency in the consolidated comparison data. Nevertheless, this excitation contained enough information through the fully non-linear Gaussian process regression to calculate the dimensional stability derivatives used to predict $\omega_{SP}$ and $\zeta_{SP}$.

\begin{table}[t]
\centering
\begin{tabular}{l ccccc}
    \hline
    Method & $C_{m_{\alpha}}$ & $C_{m_{\delta_{e}}}$ & $C_{m_{Q}}$ & $\omega_{SP}$ & $\zeta_{SP}$ \\
    \hline
    Linear Model   & $-0.285$ & $-0.525$  & $-3.240$  & $0.250$ & $0.219$ \\
    GP Estimates   & $-0.442$ & $-1.045$  & $-15.590$  & $0.317$ &$ 0.329$ \\
    \citeauthor{shepherd2010limited} Estimates \cite{shepherd2010limited} & $-0.562$   & $-1.285$    & $-12.720$    & $0.380$     & $0.290$     \\
    \hline
\end{tabular}
\vspace{0.1 in}
\caption{Comparison of dynamic parameter estimates at 0.7M and 32,000 ft. %
}
\label{tab:cm_comparison}
\end{table}

We hypothesize that we were able to reproduce high-frequency modes with low-frequency experimental content due to our imposition of a physics-based prior in the form of the Morelli mean function \cite{morelli2015global}. While the short period predicted by the A-7 model is different from the T-38C, the data ingested by our architecture refined the model's prediction towards the true value. This exemplifies the importance of using physics-informed data-based models to update aerodynamic priors based on received flight test data.

Our approach did not explicitly learn the relationship between $\omega_{SP}$ and $\bar{q}$. However, it did include $\bar{q}$ as a state variable for the GP and thus constructed a $\bar{q}$ dependence embedded within the distribution. As a result, the GP learned how implicit variations in dynamic pressure affected the forces and moments produced.

Standard flight test maneuvers like doublets provide an estimate of the short period frequency at a single dynamic pressure. Classical system identification regresses stability derivatives and produces transfer functions, but struggles to extrapolate beyond the regions where the data was collected. As a result, the test pilot must repeat the flight test techniques at a very fine grid of dynamic pressures in order to then regress across that variable. 

\begin{figure}
    \centering
    \includegraphics[width = 0.85\linewidth]{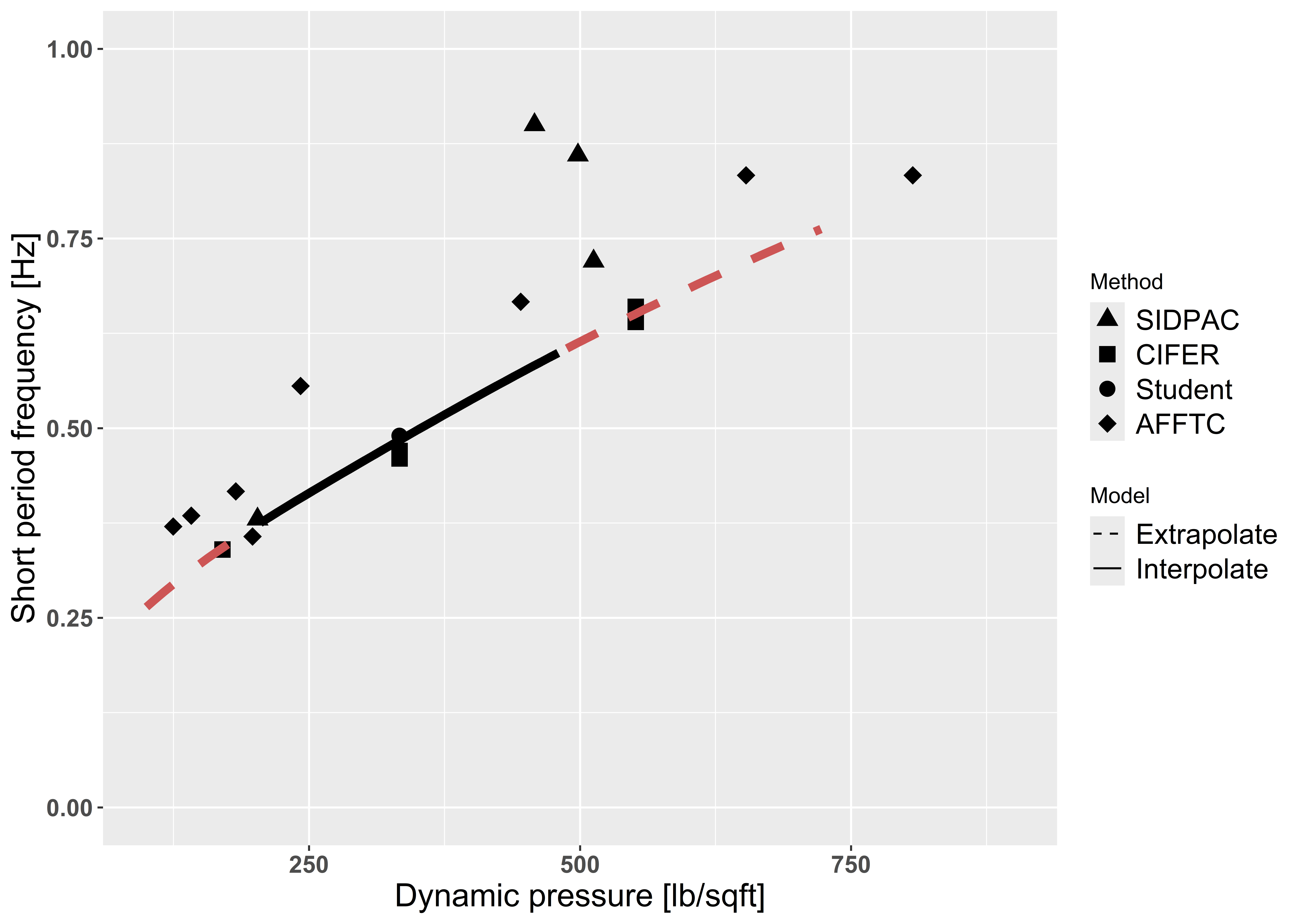}
    \caption{Short period frequency model for dependence on dynamic pressure, along with comparison data. Here, no dependency on Mach was observed.}
    \label{fig:omega-qbar}
\end{figure}

Our data-based architecture retrieves this functional dependence with no additional experimental effort by updating our aerodynamic prior across the entire flight envelope. \Cref{fig:omega-qbar} plots our result for short period frequency against the comparison data. Since the short period frequency has a weak Mach dependence, we chose a moderate Mach value of $0.7$ for this curve. Our data agrees with the lower measurements of short period frequency in the historical data.

\begin{figure}
    \centering
    \includegraphics[width = 0.85\linewidth]{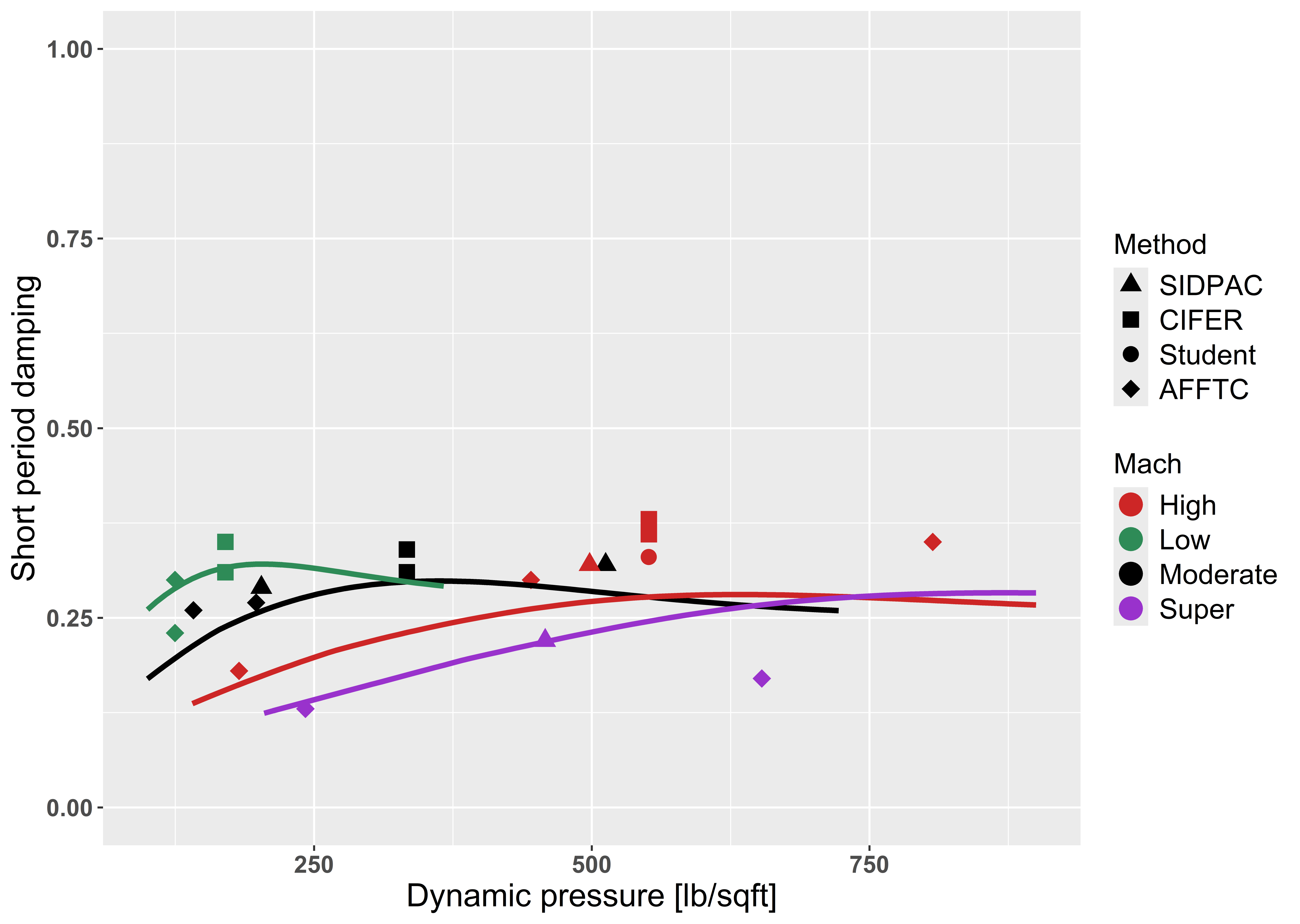}
    \caption{Short period damping model for dependence on dynamic pressure, along with comparison data. Here, different curves demonstrate the Mach dependence.}
    \label{fig:zeta-qbar}
\end{figure}

We also computed the short period damping as a function of dynamic pressure, and compare it to the historical data in \cref{fig:zeta-qbar}. In the case of damping, we observed variations in short period damping for different Mach numbers. For that reason, we calculated a family of curves for different Mach regions: ``low'' at 0.5 Mach; ``moderate'' at 0.7 Mach; ``high,'' at 0.9 Mach; and ``supersonic,'' at 1.08 Mach.  In addition to considering regions of similar dynamic pressure, we also only consider predictions of similar Mach numbers for the analysis of damping. Similar Mach values are within $\pm0.02$.

To evaluate the consistency of our model predictions against the internal variability of the historical data, we calculate the root mean square error (RMSE) across dynamic pressure regions. We group the historical data points based on dynamic pressure, considering intervals where dynamic pressures fall within $\pm 40$ \SI{}{lb/ft^2}. For each group, we compute the average RMSE for the short period frequency $ \omega_{sp} $ and damping ratio $ \zeta_{sp} $ across varying Mach numbers. This step provides a baseline measure of the inherent variability within the historical dataset.

We then calculate the RMSE of our predictions relative to the historical data, comparing them across the same dynamic pressure regions. This direct comparison allows us to assess the performance of our model in capturing the dynamics of the system compared to the baseline variability of the data. The RMSE values for both $ \omega_{sp} $ and $ \zeta_{sp} $ across different Mach number regions are summarized in \cref{tab:error_metrics}.

\begin{table}[tbh!]
    \centering
    \begin{tabular}{lcccc}
    \hline
     & \multicolumn{2}{c}{$\boldsymbol{\omega_{sp}}$} & \multicolumn{2}{c}{$\boldsymbol{\zeta_{sp}}$} \\
     \textbf{Mach} & \textbf{Prediction} & \textbf{Dataset} & \textbf{Prediction} & \textbf{Dataset}  \\
    \hline
    High      & 0.107 & 0.008  & 0.075 & 0.021 \\
    Moderate  & 0.042  & 0.011 & 0.028 & 0.011 \\
    Low       & 0.012 & 0.009  & 0.035 & 0.033 \\
    \hline
    \end{tabular}
    \vspace{0.1 in}
    \caption{Error metrics for \(\omega_{sp}\) and \(\zeta_{sp}\) across Mach regions.}
    \label{tab:error_metrics}
\end{table}

As shown in \cref{tab:error_metrics}, the RMSE values reveal that our model predictions are relatively closely aligned with the historical data in regions of low, moderate, and high Mach numbers. It is important to note that the sparsity of the historical data artificially lowers the reported RMSE values for the dataset because the mean is calculated over a limited number of data points.

\section{Conclusion}\label{sec:conclusion}
We proposed an approach to estimating aerodynamic quantities from arbitrary flight test data using Gaussian processes with physics-informed mean functions. By embedding aerodynamic principles directly within the GP framework, we developed a model capable of accurately predicting the pitching moment coefficient from diverse test conditions without the need for predefined maneuvers or tolerances on pilot execution. Our method provides accurate predictions of short-period dynamics, including the short period frequency and damping, which are essential for understanding aircraft stability characteristics across the flight envelope. The physics-informed mean function allowed us to incorporate established aerodynamic physics, enhancing both accuracy and robustness while reducing the data requirements typically associated with purely data-driven models. The proposed approach has the potential to expedite the flight test process, minimize operational costs, and enable more adaptable test strategies, offering a scalable solution for data-driven aerodynamic analysis in flight testing.

Building upon this foundation, several areas offer promising directions for further research. One area of interest is the expansion of the physics-informed GP framework to include other stability and control derivatives beyond the pitching moment coefficient through a multi-output GP approach, enabling a more comprehensive characterization of aircraft dynamics from arbitrary test data. Additionally, extending this approach to capture non-linear and unsteady aerodynamic effects—such as those encountered in high-angle-of-attack maneuvers or during rapid control inputs—would broaden its applicability in more complex flight regimes.

Further investigation could also explore real-time applications, where the model could update continuously with new flight data, enabling adaptive decision-making during flight tests. Real-time GP updates could allow test pilots and engineers to adjust test maneuvers dynamically to maximize information gain and ensure efficient data coverage. Finally, integrating the model within a feedback control framework could facilitate predictive control strategies that leverage aerodynamic insights to maintain optimal performance and safety across the flight envelope. Through these developments, the physics-informed GP model could become a cornerstone in the evolving landscape of data-driven, efficient, and safe flight testing methodologies.

\section*{Acknowledgments}
This work was partially supported by a grant from the Air Force Office of Scientific Research, Engineering and Complex Systems team, Agile Science for Test and Evaluation program office, under broad agency announcement number FA9550-23-S-0001. 

We would like to thank the United States Air Force Test Pilot School for sponsoring our staff test management project ``Have BEERKEG,'' and providing us with calibrated flight test data as well as flying additional T-38C sorties. In particular, we extend our thanks to the head of the research division, Chiawei ``FUG'' Lee, and the industrious Jessica ``Sting'' Peterson.

The views expressed in this work are those of the authors and do not necessarily reflect the official policy or position of the United States Air Force Academy, the Air Force, the Department of Defense, or the U.S. Government.

\bibliography{references}
\end{document}